\pdfoutput=1

\documentclass[11pt]{article}

\usepackage[final]{acl}

\usepackage{times}
\usepackage{latexsym}

\usepackage[T1]{fontenc}

\usepackage[utf8]{inputenc}

\usepackage{microtype}

\usepackage{inconsolata}

\usepackage{graphicx}

\usepackage{booktabs}

\usepackage{tabularx}

\usepackage[super]{nth}

\usepackage{todonotes}

\usepackage{float}
\usepackage{stfloats}

\usepackage{csquotes}

\newcommand{\narrowtext}[1]{\scalebox{0.95}[1]{#1}}

\title{Annif at the GermEval-2025 LLMs4Subjects Task: Traditional XMTC Augmented by Efficient LLMs}

\author{Osma Suominen \and Juho Inkinen \and Mona Lehtinen \\
        National Library of Finland, University of Helsinki \\ 
        \texttt{firstname.lastname@helsinki.fi}}

\begin{document}
\maketitle

\begin{abstract}
This paper presents the Annif system in the LLMs4Subjects shared task (Subtask 2) at GermEval-2025. The task required creating subject predictions for bibliographic records using large language models, with a special focus on computational efficiency. Our system, based on the Annif automated subject indexing toolkit, refines our previous system from the first LLMs4Subjects shared task, which produced excellent results. We further improved the system by using many small and efficient language models for translation and synthetic data generation and by using LLMs for ranking candidate subjects. Our system ranked 1st in the overall quantitative evaluation of and 1st in the qualitative evaluation of Subtask 2.
\end{abstract}

\section{Introduction}

Subject indexing is an important aspect of improving the discoverability of bibliographic databases and digital collections. The LLMs4Subjects Shared Task at GermEval-2025 invited teams to produce innovative systems for the application of large language models (LLMs) to the task of automated subject indexing.

We participated in the shared task \cite{dsouza-etal-2025-semeval} with an Annif\footnote{\url{https://annif.org}}-based system, as we have been developing the Annif automated subject indexing toolkit since 2017 \cite{suominen2022annif}. Our system is a more refined version of the one we used in the first LLMs4Subjects shared task at SemEval-2025 \cite{suominen2025annif}. The novel aspects of our work are 1) translation of metadata records using many different LLMs and choosing an optimal compromise between throughput and quality, 2) generation of synthetic training data using different LLMs in combination, 3) a fusion approach involving optimised weights and exponents, and 4) using a LLM for ranking candidate subjects in an ensemble.

Our system ranked first within Subtask 2 in both quantitative and qualitative evaluations. We demonstrated that our approach that combines LLMs with traditional natural language processing and machine learning remains competitive against other systems that use LLMs more heavily, and allows for different compromises between efficiency and quality - finding the best ones is important in and as of itself.

Our code, configuration files and customised data sets are available on GitHub\footnote{\url{https://github.com/NatLibFi/Annif-LLMs4Subjects-GermEval2025/}}. The models we trained are available on Hugging Face Hub\footnote{\url{https://huggingface.co/NatLibFi/Annif-LLMs4Subjects-GermEval2025-data}}.

\section{Background}

We only participated in Subtask 2: Subject Indexing. The task involved developing LLM-based systems that recommend the most relevant subjects from the GND subject vocabulary (200,035 subjects) to tag a given TIBKAT record based on its title and abstract. This is an extreme multi-label text classification (XMTC) problem.

The organisers provided bibliographic records from the TIBKAT database as a data set \cite{D_Souza_The_GermEval_2025_2025}  divided into train (n=90452), development (n=19949), and test (n=27998) subsets. Subsets were further split by document type (e.g. Article or Book) and language (German or English). GND subjects were included for the train and development records. 

As in our previous work, our system relied mainly on the traditional XMTC algorithms in Annif. However, we used LLMs to pre-process the data sets, to generate additional synthetic training data, and for the final ranking of candidate subjects.

\section{System overview}

As in our previous work, we based our system on the Annif automated subject indexing toolkit that provides a selection of XMTC algorithms. We set up base projects using three Annif backends: 1) \textbf{Omikuji}\footnote{\url{https://github.com/tomtung/omikuji}}, an implementation of a family of efficient machine learning algorithms for multilabel classification based on the idea of partitioned label trees, including Parabel \cite{prabhu2018parabel} and Bonsai \cite{khandagale2020bonsai}. We used the Bonsai-style configuration. 2) \textbf{MLLM}\footnote{\url{https://github.com/NatLibFi/Annif/wiki/Backend\%3A-MLLM}} (Maui-like Lexical Matching), a lexical algorithm for matching words and expressions in document text to terms in a subject vocabulary. It is a reimplementation of the ideas behind Maui \cite{medelyan2009human}. 3) \textbf{XTransformer}, an XMTC and ranking algorithm based on fine-tuned BERT-style Transformer models that is part of the PECOS framework \cite{yu2022pecos}. We used \textbf{FacebookAI/xlm-roberta-base} as the base model.

\label{app:llm-table}
\begin{table}[htp]
  \centering
  \footnotesize
  \setlength{\tabcolsep}{2pt} 
  \begin{tabular}{@{}llr@{}}
    \toprule
    ID & HuggingFace ID & {Size} \\
    \midrule
    A8  & \narrowtext{CohereLabs/aya-expanse-8b} & 8 \\
    E9  & \narrowtext{utter-project/EuroLLM-9B-Instruct} & 9 \\
    G4  & \narrowtext{google/gemma-3-4b-it} & 4 \\
    G12 & \narrowtext{google/gemma-3-12b-it} & 12 \\
    H3  & \narrowtext{NousResearch/Hermes-3-Llama-3.2-3B} & 3 \\
    H8  & \narrowtext{NousResearch/Hermes-3-Llama-3.1-8B} & 8 \\
    I8  & \narrowtext{ibm-granite/granite-3.3-8b-instruct} & 8 \\
    L3  & \narrowtext{meta-llama/Llama-3.2-3B-Instruct} & 3 \\
    L8  & \narrowtext{meta-llama/Llama-3.1-8B-Instruct} & 8 \\
    M8  & \narrowtext{mistralai/Ministral-8B-Instruct-2410} & 8 \\
    M12 & \narrowtext{mistralai/Mistral-Nemo-Instruct-2407} & 12 \\
    Q4  & \narrowtext{Qwen/Qwen3-4B} & 4 \\
    Q8  & \narrowtext{Qwen/Qwen3-8B} & 8 \\
    S8  & \narrowtext{VAGOsolutions/Llama-3-SauerkrautLM-8b-Instruct} & 8 \\
    S12 & \narrowtext{VAGOsolutions/SauerkrautLM-Nemo-12b-Instruct} & 12 \\
    \midrule
    G27 & \narrowtext{google/gemma-3-27b-it} & 27 \\
    M24 & \narrowtext{mistralai/Mistral-Small-3.1-24B-Instruct-2503} & 24 \\
    Q30 & \narrowtext{Qwen/Qwen3-30B-A3B} & 30 \\
    \bottomrule
  \end{tabular}
  \caption{LLMs used in our system.}
  \label{tab:llms}
\end{table}

Since a special theme for this shared task was \textbf{energy- and compute-efficient LLMs}, we specifically looked at opportunities to reduce computational costs while maintaining high performance. Thus, we mainly applied smaller, locally run (see Appendix \ref{app:llm_details}) language models and evaluated their relative performance. We selected recently published open-weight LLMs, ranging in size from 3B to 12B parameters, with support for both English and German language, including models specifically fine-tuned for improved support of German (SauerkrautLM) as well as general purpose fine-tunes (Hermes 3). We also included three models in a larger size category (24B to 30B) for comparison. The language models we used are shown in Table \ref{tab:llms}; we use the mnemonic model IDs listed in the table in the remainder of the paper.

We combined the base backends into two types of ensembles: \textit{simple ensembles} that merge subject suggestions from two or more backends by averaging their scores and a new \textit{LLM ranking ensemble}. The simple ensemble was improved from our previous work by adding support for weighting exponents (see Appendix \ref{app:ensemble-exponents} for details). The LLM ranking ensemble, in addition to averaging scores, also uses an external LLM to rank candidate subjects similar to the DNB-AI-Project system \cite{kluge2025dnb} in the first LLMs4Subjects Shared Task (see Appendix \ref{app:llm-ensemble} for details).

For evaluation during system development, we used common XMTC metrics built in to the Annif toolkit: \textit{F1@5} (F1 score calculated using the top 5 suggestions from the system) and \textit{nDCG@20} \cite{jarvelin2002}, a ranking metric calculated using the top 20 suggestions from the system. We used the Data Version Control\footnote{\url{https://dvc.org/}} tool to manage the data sets, project configurations and the training and evaluation processes.

\section{Experimental setup}

\begin{figure}[h]
  \includegraphics[width=\columnwidth]{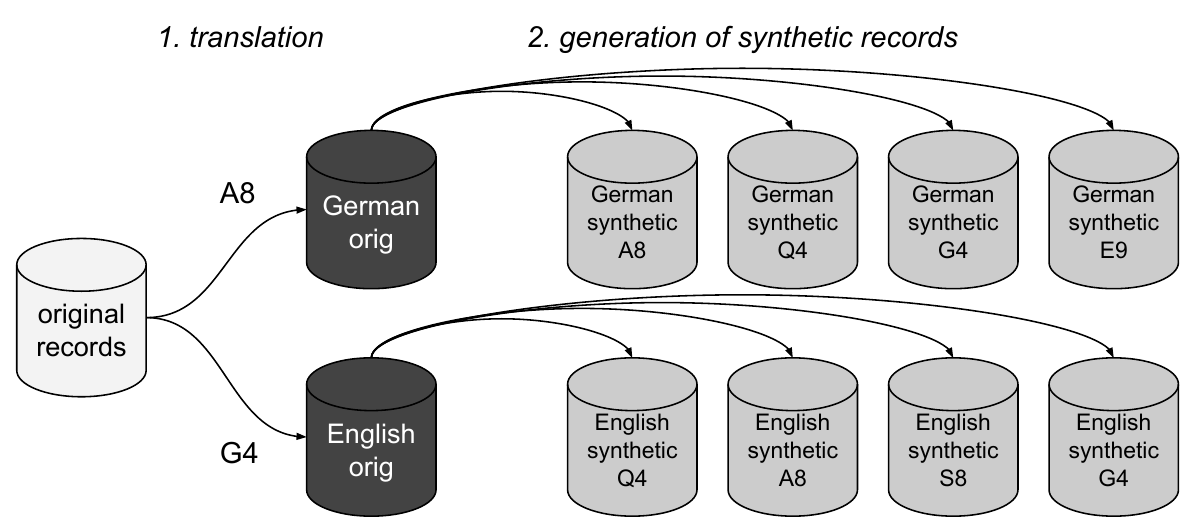}
  \caption{LLM pre-processing steps for the data sets.}
  \label{fig:preprocessing}
\end{figure}

\subsection{Translation of data sets}

For the vocabulary, we reused the bilingual English-German SKOS version of GND that was produced in our earlier work by translating the preferred terms into English using the \textbf{GPT-4o-mini} LLM.

To translate the titles and abstracts of the TIBKAT records into a consistent target language, we tested many different LLMs, aiming to find a good balance between computational efficiency and subject indexing quality. We translated each record from the train and development sets separately into German-only and English-only variants (step 1 in Figure \ref{fig:preprocessing}). We then trained Omikuji Bonsai models on the translated train records and evaluated the results against the corresponding development set. We scored the LLMs separately for English and German target languages using the formula $score = nDCG@20 + \alpha \cdot throughput$, which combines quality and efficiency into a single metric, selecting $\alpha = 0.003$ as the balancing factor. The value for the factor $\alpha$ was chosen based on experimentation and the results are shown in Figure \ref{fig:llm-translations} in Appendix \ref{app:llm-translations}. Continuing the work based on these scores, we selected the models to use for translation of TIBKAT records: G4 (Gemma 3 4B) for English and A8 (Aya 8B) for German. These were not the absolute best models in translation quality (Q30 and M24 were), but they provided the best compromise between quality and efficiency.

\subsection{Synthetic training data}

The number of training records was relatively small, so we generated additional synthetic data. For each language, we used the four top-ranking LLMs from the translation task to generate synthetic training records (titles and abstracts) in the same language. Using a one-shot approach, we presented the LLM with each of the train records (its title and abstract) at a time along with its manually assigned subject labels (GND preferred terms in either German or English, matching the language of the document). We then prompted the LLM to generate a similar record with the same set of subjects plus one additional, randomly chosen preferred term from the GND (step 2 in Figure \ref{fig:preprocessing}) that had been the subject of at least one record in the original train subset. We recognise that this process may cause seemingly unrelated subjects to appear together, but did not want to place too much constraint here; this could be the subject of further work. The result of this process was four sets of synthetic training records per language, each having the same number of records (90,452) as the original train subset.

\subsection{Base projects}

We first set up, trained and evaluated the three kinds of base projects: (Omikuji) Bonsai, MLLM and XTransformer. We aimed to maximise their evaluation scores against the development set by exploring various approaches and settings. Once satisfied with the performance of the base projects, we combined them into ensembles that were used to produce our system output.

We trained each project on the LLM-translated monolingual records from the train set (German-only or English-only, matching the project language) and tested different hyperparameters. For the Bonsai projects, we enabled bigram features using the \textit{ngram=2} setting and set a \textit{min\_df} value of 5 to filter features that occur rarely in the training data. For the XTransformer projects, we reused the model-specific hyperparameters from our previous work. The final hyperparameters can be seen in the project configuration files on GitHub\footnote{\url{https://github.com/NatLibFi/Annif-LLMs4Subjects-GermEval2025/blob/main/projects.toml}}.

\subsection{Adding synthetic data}

We evaluated the quality of the eight synthetic training record sets (2 languages \(\times\) 4 LLMs) by training Bonsai models using the original training data plus each of the synthetic sets in turn and evaluated the resulting models against the development set. The evaluation scores are shown in Table \ref{tab:eval-synthetic} in Appendix \ref{app:eval-synthetic-data}. We ranked the synthetic data sets by descending nDCG score and trained models with the original training records plus 2, 3 and 4 sets of synthetic records. The nDCG scores started to plateau after the 3rd set of synthetic data, as shown in Figure \ref{fig:synthetic-data} in Appendix \ref{app:eval-synthetic-data}. The best nDCG score was achieved using two sets of the original data (each record was used twice) and all four sets of synthetic records. We chose this setup for our final Bonsai models.

For XTransformer, we performed similar experiments. The evaluation results did not improve much after two sets of synthetic data, so we chose to use only the original records plus two sets of synthetic records for the final XTransformer models. We did not use synthetic data for MLLM because the original records were sufficient.

The final evaluation scores for the base projects are shown in Table \ref{tab:eval_base_projects} in Appendix \ref{app:eval-base-projects}. The Bonsai projects achieved the best scores in all cases, followed by XTransformer and MLLM.

\begin{table*}[t]
  \centering
  \footnotesize
  \begin{tabular}{llcccccccc}
    \toprule
    \multicolumn{3}{c}{} & \multicolumn{2}{c}{development set / de} & \multicolumn{2}{c}{development set / en} & \multicolumn{2}{c}{test set / de+en} \\
    \cmidrule(lr){4-5} \cmidrule(lr){6-7} \cmidrule(lr){8-9}
    System & Ensemble type &  & F1@5 & nDCG@20 & F1@5 & nDCG@20 & F1@5 & nDCG@20 & Rank \\
    \midrule
    Annif (ours) & BM simple  &  & 0.3344 & 0.5923 & 0.3345 & 0.6101 & 0.3200 & 0.5496 \\
                 & BMX simple &  & 0.3386 & 0.6012 & 0.3479 & 0.6162 & 0.3200 & 0.5507 \\
                 & BMX LLM Q4 &  & 0.3507 & 0.6259 & 0.3547 & 0.6329 & 0.3251 & 0.5631 \\
                 & BMX LLM M24 & & \textbf{0.3587} & \textbf{0.6375} & \textbf{0.3600} & \textbf{0.6412} & \textbf{0.3288} & \textbf{0.5697} & \nth{1} \\
    \midrule
    Annif (ours)             &    & & & & & & \textbf{0.3288} & \textbf{0.5697} & \nth{1} \\
    DNB-AI-Project &    & & & & & & 0.2466 & 0.4180 & \nth{2} \\
    ubffm            &    & & & & & & 0.0742 & 0.1465 & \nth{3} \\
    \bottomrule
  \end{tabular}
  \caption{Quantitative evaluation results for the ensemble projects measured against the development and test sets. Other participating systems included for comparison. Development set results are split by language of the Annif project (de/en), while the test set results were produced using bilingual predictions.}
  \label{tab:eval_ensemble_projects}
\end{table*}

\subsection{Ensemble projects}

\begin{figure}[htp]
  \includegraphics[width=\columnwidth]{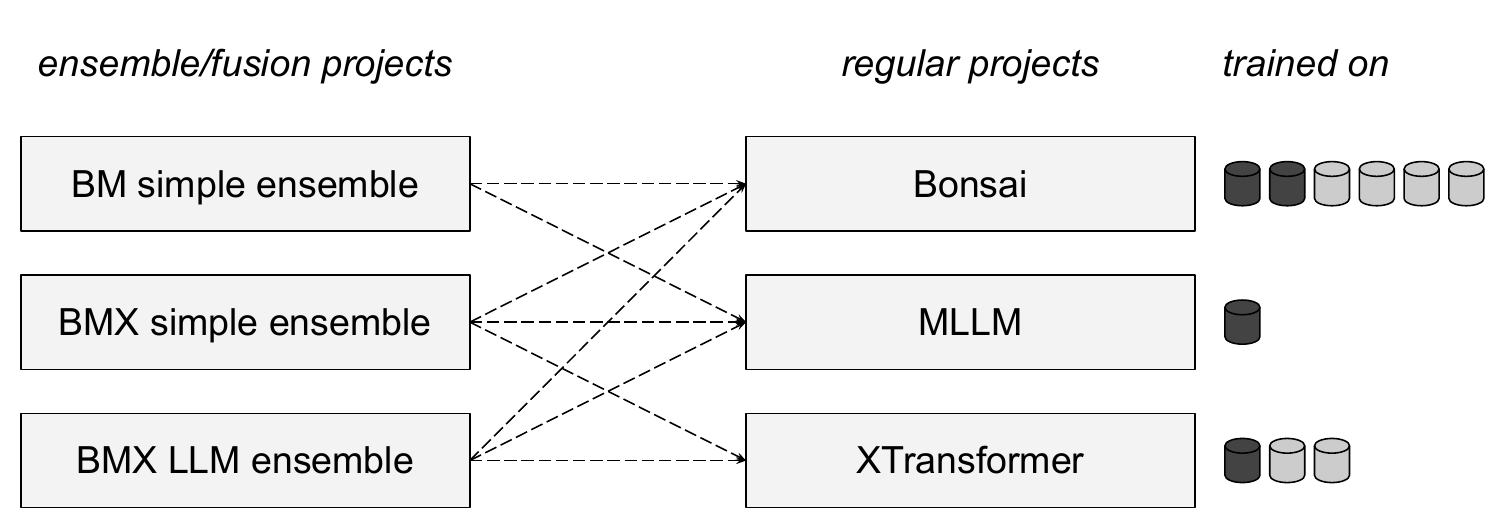}
  \caption{Annif base projects and ensembles.}
  \label{fig:projects}
\end{figure}

We set up three kinds of ensemble projects that combined the base projects in different ways: a "BM" simple ensemble combining Bonsai and MLLM, and two "BMX" ensembles (simple + LLM ranking) that combine all three base projects (see Figure \ref{fig:projects}). We then tuned the simple ensembles using the \texttt{annif hyperopt} command to try 400 different combinations of weights and exponents per project, choosing hyperparameters that maximised the nDCG scores against the development set (see Table \ref{tab:ensemble_weights} in Appendix \ref{app:ensemble}).

In the BM ensembles, the Bonsai model contributed 85--90\% while MLLM had a minor role. In the BMX ensembles, the Bonsai model contributed the most for English (46\%), while XTransformer was the most important model for German (44\%). The weighting exponents ranged from 0.63 to 1.17. We reused the optimised weights and exponents of the BMX simple ensembles also for the corresponding BMX LLM ensembles.
The weighting exponents are intended to account for the differing scoring scales of the backends, as the relative relevance of the algorithms can vary depending on the score range.
We set up two variants of the LLM ensembles: 1) using the smaller Q4 LLM and 2) using the larger M24 LLM. The results of evaluating the ensembles against the development set are shown in Table \ref{tab:eval_ensemble_projects}.

Teams were allowed to submit up to 10 different system outputs (up to 20 subject predictions per test set record) to the Codabench system\footnote{\url{https://www.codabench.org/competitions/8373/}} used for evaluation. We translated the test records to produce German-only and English-only versions of each record using the same LLM choices as for the train and development records. For each of the ensemble types, we first produced 100 predictions each using the German and English versions of the ensembles, then merged the predictions by summing the scores of the predicted subjects and choosing the top 20 subjects by score. We submitted these bilingual predictions.

\section{Results}

The main quantitative metric used for ranking systems was nDCG@20, but other metrics such as precision, recall, F1 score and R-precision were also provided by the Codabench system. The bilingual predictions of our LLM ranking ensemble using the M24 LLM ranked \nth{1} with a nDCG@20 score of 0.5697 (see Table \ref{tab:eval_ensemble_projects}). 

\begin{table}[H]
    \centering
    \footnotesize
    \setlength{\tabcolsep}{4pt} 
    \begin{tabular}{lccc}
        \toprule
        System & nDCG@20 C1 & nDCG@20 C2 & Rank \\ 
        \midrule
        Annif (ours)& \textbf{0.8787} & \textbf{0.7699} & \nth{1} \\
        DNB-AI-Project & 0.6545 & 0.5837 & \nth{2} \\
        ubffm & 0.3454 & 0.2471 & \nth{3} \\
        \bottomrule
    \end{tabular}
    \caption{Qualitative evaluation results for systems in cases 1 (C1) and 2 (C2)}
    \label{tab:qualitative_eval}
\end{table}

Subject librarians conducted a qualitative evaluation as well: five different subject classification categories were considered, each containing 10 records, within which the top 20 GND codes for each submission were assessed as correct (Y), technically correct but irrelevant (I), or incorrect (N or blank). Two types qualitative results were calculated: one considering both Y and I as correct (Case 1), and another considering only Y as correct (Case 2). The results are detailed in Table \ref{tab:qualitative_eval}. Annif was ranked first in both cases with the overall nDCG@20 being 0.8787 in the first and 0.7699 in the second case.

\newpage

\section{Conclusions}

We experimented with different local LLMs and found that some models, including A8 (Aya 8B), G4 (Gemma3 4B) and Q4 (Qwen3 4B) have a particularly good balance of quality and efficiency. By using LLM-generated synthetic training data, we could improve the results of traditional XMTC models: Bonsai improved a lot, while the improvements for XTransformer were less dramatic. The new LLM ranking ensemble approach improved the nDCG scores of topic suggestions by 0.01-0.03 compared to simple ensembles. The four different ensembles allow different compromises to be made between quality and computational efficiency, with the best scores achieved using a larger M24 LLM and somewhat lower scores for the smaller Q4 LLM and the simpler ensembles that do not require an LLM at all.

\section*{Acknowledgments}

The authors wish to thank the Finnish Computing Competence Infrastructure (FCCI) for supporting this project with computational and data storage resources. 

\bibliography{custom}

\appendix

\section{Ensemble with weighting exponents}
\label{app:ensemble-exponents}

Annif supports a simple ensemble approach that performs weighted averaging of prediction scores from multiple projects. Automated hyperparameter optimisation can be used to discover the weight values that maximise the evaluation results measured against a development set.

For this task, we enhanced the Annif simple ensemble by adding support for weighting exponents. The concept-specific prediction scores of the simple ensemble are calculated using the formula
\[
f = \sum_{k=1}^{n} w_k \cdot x_k^{p_k}
\]
where $f$ is the final combined prediction score, $x_k$ is the prediction score from the $k$-th project, $w_k$ is the weight coefficient for the $k$-th project (weights add up to 1.0), $p_k$ is the power exponent for the $k$-th project and $n$ is the total number of projects in the ensemble. Along with the weights, the optimal exponents can be found using hyperparameter optimisation.

\section{LLM ranking ensemble}
\label{app:llm-ensemble}

We implemented a new type of ensemble in Annif based on the idea of an LLM that ranks candidate subjects based on how well they describe a given text. As in the simple ensemble, the subject predictions from multiple projects are first combined using configurable weights and exponents. The top $K$ combined predictions (we chose $K=100$) are then given to an LLM along with the document text, and the LLM is prompted to respond with relevance scores (from 0 to 100) for each candidate subject. The final subject predictions of the LLM ranking ensemble are calculated using the formula
\[
f = w \cdot r^p + (1-w)x 
\]
where $x$ is the combined prediction from the source projects (calculated as in the simple ensemble above), $r$ is the relevance score given by the LLM (rescaled to the interval $[0.0, 1.0]$), $w$ is the weight (0.0 to 1.0) and $p$ is the weighting exponent given to the LLM prediction. The optimal weights and exponents for the LLM ranking can be found using hyperparameter optimisation.

\section{Details about LLM usage}
\label{app:llm_details}

We used the vLLM\footnote{\url{https://docs.vllm.ai}} inference engine to translate and synthesise records and to rank candidate subjects. For the processing, we used a single NVIDIA A100 GPU with 80GB VRAM from the University of Helsinki HPC cluster Turso.

\subsection{Prompt templates}

In the following prompt templates, the system prompt is set in \textit{italics}. Tags in the template were replaced with relevant information from the records. The prompt templates are the same as in our earlier work \cite{suominen2025annif}, except for the ranking prompt, which is new to this system.

\subsubsection{Record translation}

This prompt was used to translate records:

\begingroup
\setlength{\parindent}{0pt}
\setlength{\parskip}{4pt} 
\raggedright
\small
\rule{\columnwidth}{0.5pt}
\textit{You are a professional translator specialized in translating bibliographic metadata.}

Your task is to ensure that the given document title and description are in <LANGUAGE> language, translating the text if necessary. If the text is already in <LANGUAGE>, do not change or summarize it, keep it all as it is.

Respond with only the text, nothing else.

Give this title and description in <LANGUAGE>:

<TITLE>

<DESCRIPTION> \\
\rule{\columnwidth}{0.5pt}
\endgroup

\subsubsection{Record synthesis}

This prompt was used to synthesise new records:

\begingroup
\setlength{\parindent}{0pt}
\setlength{\parskip}{4pt} 
\raggedright
\small
\rule{\columnwidth}{0.5pt}
\textit{You are a professional metadata manager.}

Your task is to create new bibliographic metadata: document titles and descriptions.

Here is an example document title and description in <LANGUAGE> with the following subject keywords: <OLD\_KEYWORDS>

<TITLE\_DESC>

Generate a new document title and description in <LANGUAGE>. Respond with only the title and description, nothing else. Create a new title and description that match the following subject keywords: <NEW\_KEYWORDS> \\
\rule{\columnwidth}{0.5pt}
\endgroup

\subsubsection{Ranking of candidate concepts}

This prompt was used to rank candidate concepts:

\begingroup
\setlength{\parindent}{0pt}
\setlength{\parskip}{4pt} 
\raggedright
\small
\rule{\columnwidth}{0.5pt}
\textit{You will be given text and a list of keywords to describe it. Your task is
to score the keywords with a value between 0 and 100. The score value
should depend on how well the keyword represents the text: a perfect
keyword should have score 100 and completely unrelated keyword score
0. You must output JSON with keywords as field names and add their scores
as field values.
There must be the same number of objects in the JSON as there are lines in
the intput keyword list; do not skip scoring any keywords. \\ }
\rule{\columnwidth}{0.5pt}
\endgroup

\section{LLMs for translating metadata}
\label{app:llm-translations}

\begin{figure*}[ht]
  \includegraphics[width=\textwidth]{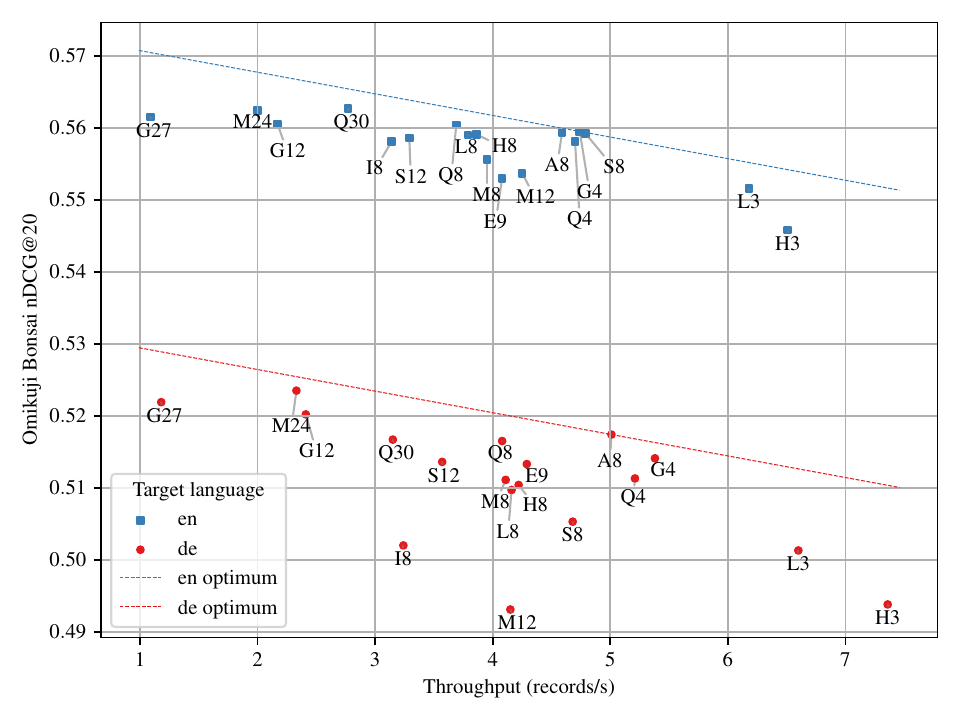}
  \caption{Evaluation of different LLMs for translating metadata records into English and German. The lines indicate the maximum value for the combined quality and efficiency score; the best models are located near the lines.}
  \label{fig:llm-translations}
\end{figure*}

The evaluation results of different LLMs for translating metadata are shown in Figure \ref{fig:llm-translations}.

\section{Evaluation results for synthetic data}
\label{app:eval-synthetic-data}
\begin{table}[htp]
    \centering
    \begin{tabular}{clc}
        \toprule
        Lang & Train data & nDCG@20  \\
        \midrule
        de & orig    & 0.5122 \\
           & orig+A8 & \textbf{0.5513} \\
           & orig+Q4 & 0.5492 \\
           & orig+G4 & 0.5488 \\
           & orig+E9 & 0.5469 \\
        \midrule
        en & orig    & 0.5595 \\
           & orig+Q4 & \textbf{0.5896} \\
           & orig+A8 & 0.5889 \\
           & orig+S8 & 0.5845 \\
           & orig+G4 & 0.5845 \\        
        \bottomrule
    \end{tabular}
    \caption{Evaluation scores when evaluating Omikuji Bonsai trained using various combinations of original and synthetic data against the development set.}
    \label{tab:eval-synthetic}
\end{table}

\begin{figure*}[htp]
  \includegraphics[width=\textwidth]{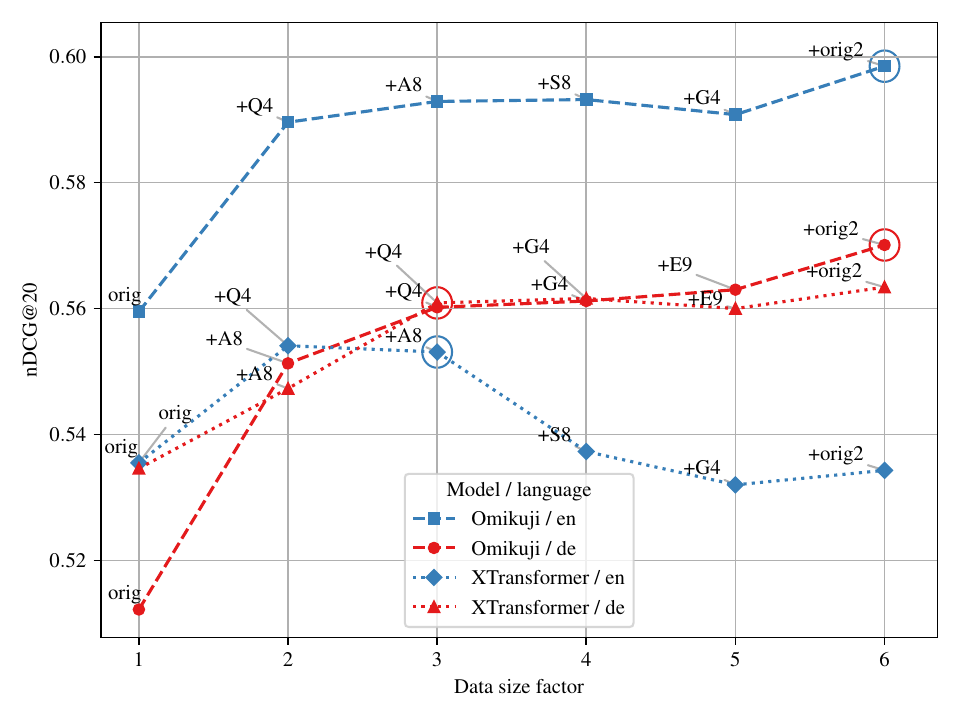}
  \caption{Effect of adding synthetic training data for Omikuji Bonsai and XTransformer models. The models were evaluated against the development set. The circles indicate our final choice of synthetic data: four sets of synthetic data and a doubling of the original records for Omikuji Bonsai, and two sets of synthetic data for XTransformer.}
  \label{fig:synthetic-data}
\end{figure*}

The evaluation results of different LLMs for producing synthetic records are shown in Table \ref{tab:eval-synthetic}. All the tested models produced similar levels of improvement in terms of nDCG score. Figure \ref{fig:synthetic-data} shows the improvement in nDCG scores after adding multiple sets of synthetic data to the original data.

\section{Results for base projects}
\label{app:eval-base-projects}

\begin{table}[htp]
    \centering
    \begin{tabular}{cccc}
        \toprule
        Backend & Lang & F1@5 & nDCG@20 \\
        \midrule
        Bonsai &        de & \textbf{0.3229} & \textbf{0.5701} \\
        &               en & \textbf{0.3401} & \textbf{0.5985} \\
        MLLM &          de & 0.2487 & 0.4437 \\
        &               en & 0.2304 & 0.4180 \\
        XTransformer &  de & 0.3189 & 0.5609 \\
        &               en & 0.3122 & 0.5517 \\
        \bottomrule
    \end{tabular}
    \caption{Evaluation scores of the base projects measured against the development set. The best scores for each language have been set in \textbf{bold}.}
    \label{tab:eval_base_projects}
\end{table}

The evaluation results of the final base projects are shown in Table \ref{tab:eval_base_projects}.

\section{Ensemble overview, weights and exponents}
\label{app:ensemble}

\begin{table}[htp]
    \centering
    \small
    \setlength{\tabcolsep}{2pt} 
    \begin{tabular*}{\columnwidth}{@{\extracolsep{\fill}}cccc}
        \toprule
        Type & Bonsai & MLLM & XTransformer \\
        \midrule
        BM/de & 0.8523:1.1119 & 0.1477:1.1611 & - \\
        BM/en & 0.9015:1.0342 & 0.0985:0.8483 & - \\
        \midrule
        BMX/de & 0.3823:1.0457 & 0.1734:1.1675 & 0.4443:1.0794 \\
        BMX/en & 0.4608:0.6855 & 0.2403:0.8567 & 0.2989:0.6290 \\
        \bottomrule
    \end{tabular*}
    \caption{Optimised weights and exponents (given as $weight:exponent$) of the base projects in the BM and BMX ensembles.}
    \label{tab:ensemble_weights}
\end{table}

An overview of the configurations of the BM and BMX ensembles with optimised weights and exponents is shown in Table \ref{tab:ensemble_weights}. For the LLM ranking ensemble, we used an LLM weight of 0.1592 (exponent 7.898) for English and 0.0921 (exponent 9.328) for German. Thus, the LLM ranking had only a relatively small effect on individual prediction scores, but overall output quality improved, as shown in the evaluation results.

\end{document}